\pdfoutput=1

\documentclass[11pt]{article}

\usepackage{acl}

\usepackage{times}
\usepackage{latexsym}

\usepackage[T1]{fontenc}

\usepackage[utf8]{inputenc}

\usepackage{microtype}
\usepackage{graphicx}
\usepackage{booktabs}       
\usepackage{amsfonts}       
\usepackage{nicefrac}
\usepackage{latexsym}
\usepackage{multirow}
\usepackage{amsmath}
\usepackage{arydshln}
\usepackage{subfigure}
\usepackage{colortbl}
\usepackage{hyperref}
\title{RetrievalSum: A Retrieval Enhanced Framework for Abstractive Summarization}

\author{Chenxin An$^1$,
Ming Zhong$^2$,
Zhichao Geng$^1$,
Jianqiang Yang$^1$,
Xipeng Qiu\thanks{\ \  Corresponding author.}$^1$\\
$^1$Shanghai Key Laboratory of Intelligent Information Processing, Fudan University\\
    $^1$School of Computer Science, Fudan University\\
    $^2$University of Illinois at Urbana-Champaign \\
\texttt{\{cxan20, zcgeng20, jqyang20, xpqiu\}@fudan.edu.cn, mingz5@illinois.edu}
}

\definecolor{mygray}{gray}{0.9}

\begin{document}

\maketitle

\begin{abstract}
Existing summarization systems mostly generate summaries purely relying on the content of the source document. However, even for humans, we usually need some references or exemplars to help us fully understand the source document and write summaries in a particular format. But how to retrieve the high-quality exemplars and incorporate them into summarization systems is still challenging and worth exploring. In this paper, we propose \textsc{RetrievalSum}, a novel retrieval enhanced abstractive summarization framework consisting of a dense \textit{Retriever} and a \textit{Summarizer}. Retrieved exemplars are not only additional knowledge  but also  guidance of the writing style of a specific corpus. We validate our method on a wide range of summarization datasets across multiple domains and two backbone models: BERT and BART. Results show that our retrieval enhanced framework can greatly improve baselines with different powerful pre-trained models by $1.38\sim4.66$ in ROUGE-1 score. 
\end{abstract}
\section{Introduction}
\begin{figure}[t]
    \centering
    \includegraphics[width=0.95\linewidth]{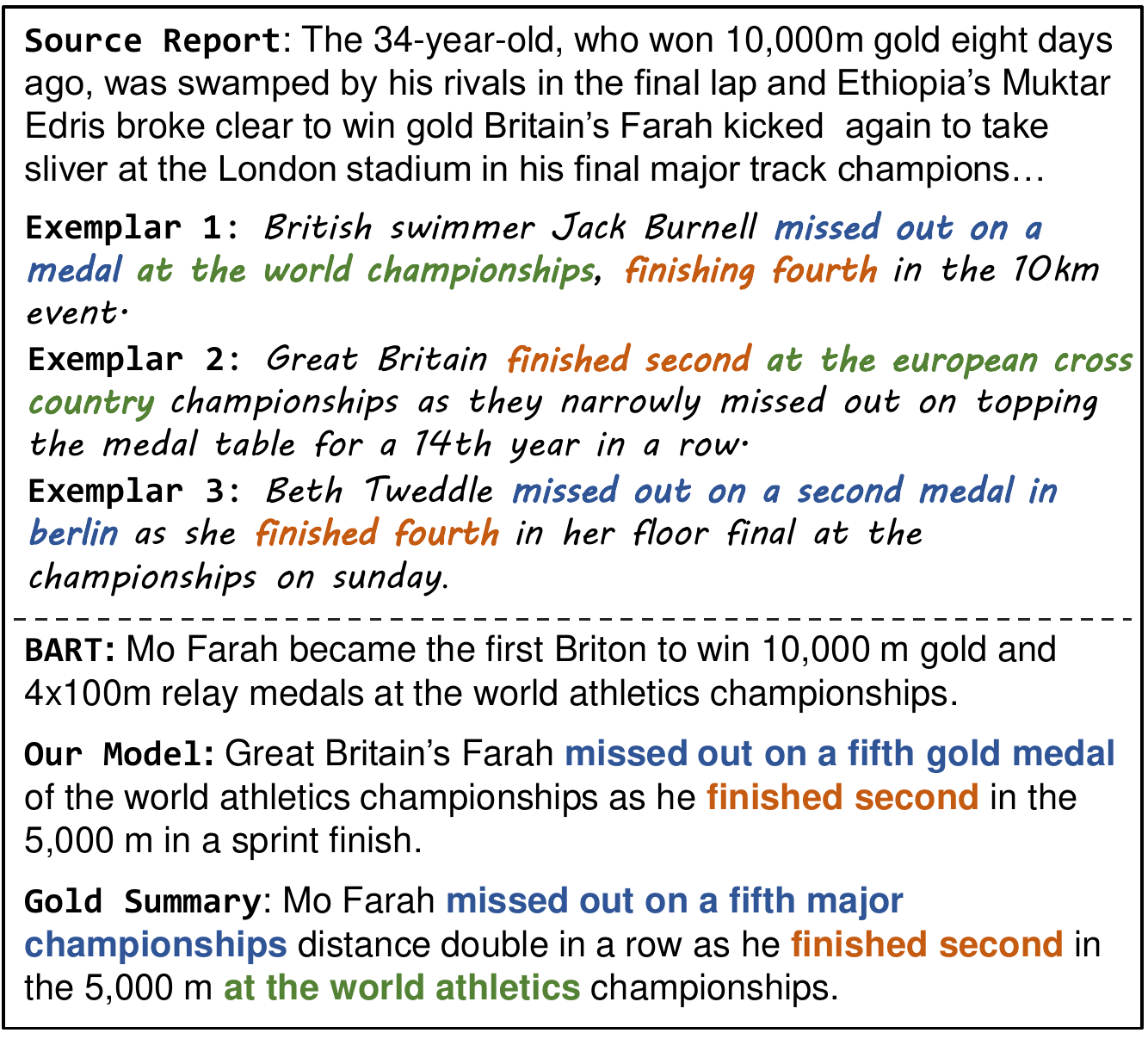}
    \caption{An case study from XSum dataset where the source document is a sports news. We retrieve three exemplars from the training corpus which are summaries of other sports events. The blue, green, orange text represents the key information required in sport news: the result of the sports event, the name of the event and the ranking of the athletes. Our model can generate more relevant and correctly formatted summaries than BART.}
    \label{fig:case_study}
\end{figure}

Text summarization aims to compress a document automatically into a shorter text while retaining the main idea~\cite{nallapati2017summarunner,zhong2019searching}. Recently, the application of pre-trained language models in text summarization has attracted increasing attention and achieved state-of-the-art~\cite{liu2019text, lewis2020bart} performance.

However, most existing summarization systems purely rely on the source document, which makes the model
suffer from problems such as \textit{content deviation} and \textit{inconsistent writing style}~\cite{jin2020semsum,cao2018retrieve, saito2020abstractive} at the same time. Regarding the first issue, recent work~\cite{li2018guiding, dou2020gsum} reveals that even for the powerful pre-trained models, a single document is not enough for the model to fully understand its content and generate a qualified summary. Thus they attempt to explore external guidance such as highlighted sentences, keywords to assist the generation process. On the other hand, in many application scenarios such as government reports and scientific papers, the abstracts are expected to be written in a formal and specific style, which is a capability that general-purpose abstractive models do not have.

To solve the above issues, a natural way is to introduce some domain-specific writing templates to teach the model to generate text that meets the requirements. We can directly instantiate the \textit{exemplars} here as the reference summaries in the training corpus with following motivations: 1) gold summaries from the training corpus are high-quality and easy-to-obtain resources; 2) reference summaries contain highly condensed domain-specific knowledge; 3) learning from the writing style in the same domain is essential to write a consistent and coherent summary. Figure~\ref{fig:case_study} is a case study from XSum dataset~\cite{narayan2018don}, we show three retrieved exemplars from other sports news in the training set of XSum. Without the help of exemplars, even the state-of-the-art model BART~\cite{lewis2020bart} generates summary with deviated content and incorrect format.

In this paper, we propose \textsc{RetrievalSum}, a retrieval augmented framework for abstractive summarization.  Introducing retrieved exemplars into summarization can bring not only guidance of writing format but also additional background knowledge. For example, the meeting summarization dataset AMI~\cite{mccowan2005ami} about designing remote controls, important decisions made in previous meeting are also useful information to summarize the current meeting. Our model consists of two modules: \textit{Retriever} and \textit{Summarizer}. Given a query document, the \textit{Retriever} is  asked to retrieve the most related exemplars from the knowledge base. Previous work~\cite{cao2018retrieve, dou2020gsum}  always use the term-based systems such as TF-IDF, BM25, but we argue that it may not a good solution to directly perform lexical matching between the query document and candidate exemplars. Motivated by the success of dense passage retriever (DPR)~\cite{karpukhin2020dense}, we design our exemplars retrieval as a semantic matching process. The second module \textit{Summarizer} generates the summary not only relying on source document but also the retrieved exemplars. We hope the retrieved exemplars can be easily integrated into any summarization models without requiring modification of their inner structure. With a series of experiments, our final solution contains two mechanisms: group alignment and ROUGE credit~\cite{an2021enhancing}. To force our model learn the structure of the exemplars, we perform group alignment on each sentence in the exemplars and the generated summary during training. ROUGE Credit is to encourage the model to keep beams with high style consistency with the retrieved exemplars during beam search.


We summarize our contributions as follows:

1) We introduce retrieval techniques into abstractive summarization and propose an effective  exemplar retriever to retrieve a set of candidate exemplars for given a query document which greatly outperforms the traditional sparse vector space models in the downstream summarization task.

2) We explore two experimentally powerful mechanisms to incorporate retrieved exemplars into abstractive summarization systems. Meanwhile, our approach are easy to transfer to most summarization model because we do not require modifying the inner structure of backbone models. 

3) Experimental results show that retrieved exemplars can substantially improve baseline models initialized with different pre-trained models. Human evaluation also shows that our generation results achieve close-to-human performance in terms of writing style and concrete expressions.\\


\section{Related Work}
\subsection{Summarization with Guidance}
Previous work show that generate summarization purely relies on document is not enough. An intuitive idea is to extract some keywords~\cite{li2018guiding} as a guidance and incorporate them into the decoding stage.  As an extension for the keywords based guidance, \citet{jin2020semsum, DBLP:conf/acl/HuangWW20} first extract relation triples and entities from the input document and then encode these relations by Graph Neural Networks. A concurrent related work is GSum~\cite{dou2020gsum} which is a general framework making use of various type of guidance. In the scientific domain, it is more challenging to understand the paper. Thus \citet{an2021enhancing} propose a summarization task with the guidance signals from citation graph.

\subsection{Template-based Methods}
Early works~\cite{oya2014template,cao2018retrieve, gao2019write} have focused on utilizing task-specific templates to make the generation results more informative before large pre-trained models successfully applied in generation tasks. \citet{oya2014template} construct templates for meeting using multi-sentence fusion algorithm. \citet{cao2018retrieve} form a retrieve-rerank-rewrite pipeline to facilitate templates. \citet{gao2019write} find that court judgments always required a particular style thus they manually build prototype document-summary pairs and employ explicit  prototype editing. Compared with these methods, on the one hand, our framework do not require human involvement and appliable to most pretrained summarzation models. Due to our exemplars are not human-made, only one exemplar will lead to error accumulation hence we provide the pre-trained backbone multiple exemplars and let the model implicitly learn which exemplar to refer to by the seq2seq loss. On the other hand, instead of using the traditional retrieve techniques like previous work~\cite{cao2018retrieve, dou2020gsum}, our dense retriever is trained with contrastive learning whose negative samples are exemplars with high lexical matching score but actually unhelpful for summarizing this document.

\section{Method}
In this section, we describe our summarization pipeline in detail.

\subsection{Problem Formalization}
Text summarization is usually considered as a sequence-to-sequence task. Given a document $D$, the goal of the model is to generate a shorter summary $Y$. We introduce a knowledge base $\mathcal{K}$ on this basis, which can be the current dataset or an additional corpus. Before feeding the document $X$ into the model, we firstly retrieve some semantically-close exemplars $\mathcal{E}$ $ = ({E}_{1}, {E}_{2},\ldots,{E}_{|\mathcal{E}|})$ from $\mathcal{K}$. Here, $E$ can either be a document or a target summary. The objective of our framework is to model the conditional distribution $P(Y|X; \mathcal{E})$. We instantiate $\mathcal{K}$ as the training set, that is, $\mathcal{K} = \{(X_1, Y_1), \ldots,(X_{|\mathcal{K}|}, Y_{|\mathcal{K}|})\}$ consisting of $|\mathcal{K}|$ document-summary pairs. In this work,  we instantiate $\mathcal{E}$ as a set of target summaries, $\mathcal{E}$ $ = ({Y}_{1}, \ldots,{Y}_{|\mathcal{E}|})$.

\begin{figure}[t]
    \centering
    \includegraphics[width=0.9\linewidth]{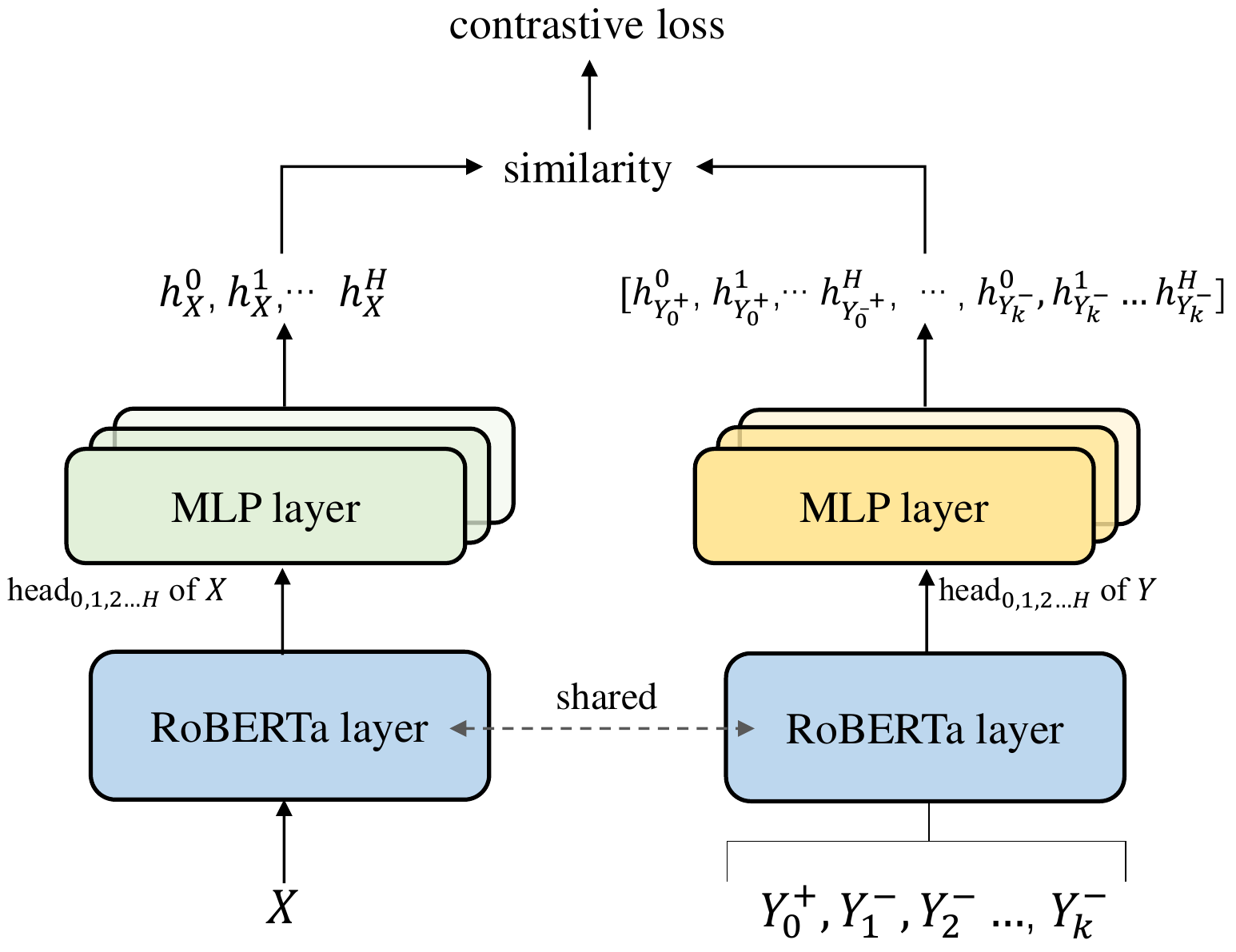}
        \caption{Overview of retriever. $X$ is the input document and $Y_0^+, Y_1^- ,\ldots, Y_k^-$ are its corresponding candidates exemplars from $\mathcal{K}$ where $Y_i^+$ / $Y_i^-$ means the $i$-th positive/negative candidates. Whether a candidate $Y_i$ is positive or negative is determined by the ROUGE score between  $Y_i$ and the target summary of X. }
    \label{fig:contracstive}
\end{figure}

\subsection{Retriever}
Different from the existing summarization system, we introduce a dense \textit{retriever} to retrieve some useful exemplars to help the entire summarization process. Formally, given a document $X$, retriever need to find a set of reference summaries $\mathcal{E}$ from $\mathcal{K}$, where $\mathcal{E}$ should have the highest similarity scores with $X$. Instead of simply employing traditional IR algorithms,  we assume the document and its summary always convey the same meaning thus they should be close in semantic space and useful exemplars should also be close to its gold summary.  Therefore, we conceptualize this retrieval procedure as a semantic text matching problem.
We utilize a Siamese Network architecture~\cite{DBLP:conf/acl/ZhongLCWQH20} to match the document $X$ and the candidate exemplars. As is shown in Figure~\ref{fig:contracstive}, the current document $X$ and some candidate exemplars $Y_0, Y_1,\ldots, Y_k$ are input into a shared encoder layer and a non-shared MLP layer to get their respective representations through the special token `[CLS]'. We use the ROUGE score of $Y_i$ and $Y$ to determine whether the current candidate is a positive or negative sample, where the positive sample should have a higher similarity score with $X$. 
\paragraph{Coarse-grained Ranking}
Matching source document to all candidates in $\mathcal{K}$ will result in out of memory problem because the size of training set is usually excessively large. Thus it is necessary to perform a coarse-grained ranking first. Concretely, we use a BertExt model~\cite{liu2019text} to extract the salience sentences in document $X$. Then we calculate the ROUGE score between the extractive sentences in $X$ and all the candidates. In the experiments of this paper, after coarse-grained ranking, we reserve 100 candidate examplars with the highest scores for each document.
\paragraph{Contrastive Loss} Instead of using the whole corpus $\mathcal{K}$, we sample a subset for training our retriever to save time. 
We denote the candidates set after coarse-gained ranking as $\mathcal{C}$. We sort all the candidate exemplars in $\mathcal{C}$ by the ROUGE score with reference summary $Y$, and select top 8 candidates with the highest score as positive samples in the experiments. We choose cosine similarity score as our similarity score and adopt a form of a contrastive loss function proposed in MoCo~\cite{DBLP:conf/cvpr/He0WXG20}:
\begin{equation}
\label{eq:nceloss}
\mathcal{L} = \sum_{j=1}^{|\mathcal{C}^+|}-\mathrm{log}\frac{\mathrm{exp}(s_j^+/\tau)}{\mathrm{exp}(s_j^+/\tau) + \sum_{i}\mathrm{exp}(s_i^-/\tau)},
\end{equation}
where $\tau$ is a temperature hyper-parameter, $s_j^+ =\mathrm{cos}(\mathbf{h}_X, \mathbf{h}_{Y_j^+})$ and $s_i^- =\mathrm{cos}(\mathbf{h}_X, \mathbf{h}_{Y_i^-})$ is the cosine score of positive and negative samples, $|\mathcal{C}^+|$ is the size of positive sample set and $\mathbf{h}_X$, $\mathbf{h}_{Y_j^+}$, $\mathbf{h}_{Y_i^-}$ are the representation of the document $X$, the $j$-th positive candidate, $i$-th negative candidate, respectively. We extend the computation of cosine similarity to multi-head.  The procedure of computing multi-head similarity score between $X$ and candidates is similar with multi-head attention in Transformer.
The similarity score calculated from the $k$-th head is:
\begin{align}
\label{pos_score-m}
 (s_j^+)^k &= \mathrm{cos}(\mathbf{h}_X^k, \mathbf{h}_{Y_j^+}^k), \\
\label{neg_score-m}
 (s_i^-)^k &= \mathrm{cos}(\mathbf{h}_X^k, \mathbf{h}_{Y_i^-}^k).
\end{align}

The final contrastive loss is derived by summing up the loss over all heads:
\begin{small}
\begin{equation}
\label{eq:nceloss-m}
 \mathcal{L} = \sum_{k=1}^H\sum_{j=1}^{|\mathcal{C}^+|}-\mathrm{log}\frac{\mathrm{exp}((s_j^+)^k/\tau)}{\mathrm{exp}((s_j^+)^k/\tau) + \sum_{i}\mathrm{exp}((s_i^-)^k/\tau)},
\end{equation}
\end{small}
where $H$ is the number of heads. We obtain the final exemplars set $\mathcal{E}$ by the voting result of all the $H$ heads. Concretely, each head will select its top $e$ best candidates based on cosine similarity during inference. Among the selected candidates from all head, we further choose the top $e$ most frequently occurring  summaries to form the final exemplars set $\mathcal{E}$.

\begin{figure*}[t]
    \centering
    \includegraphics[width=0.98\linewidth]{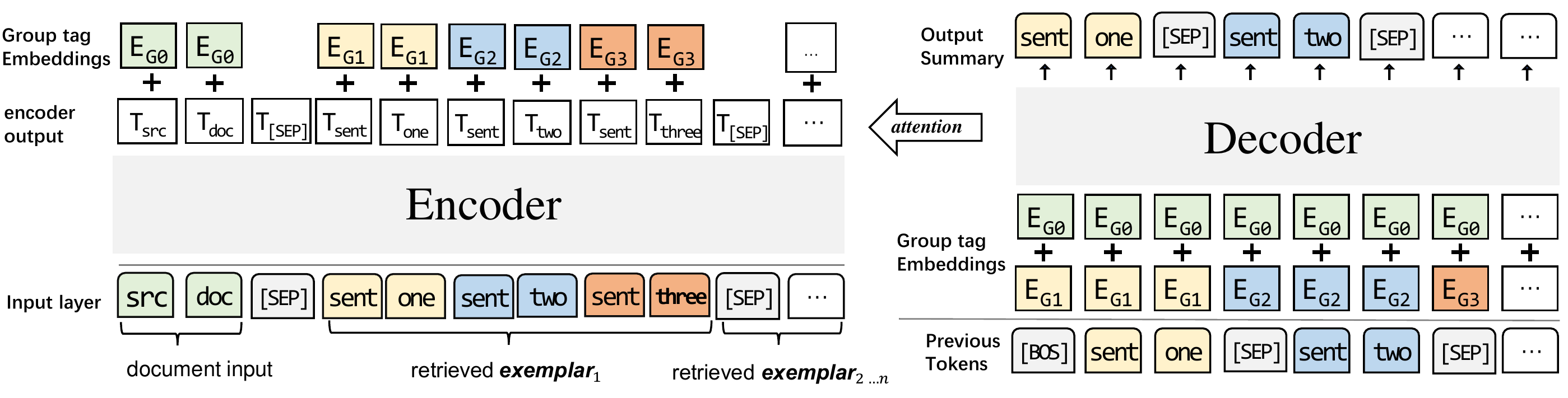}
    \vspace{-10pt}
    \caption{\label{fig:architecture} Architecture of the summarizer with retrieved exemplars. Except for the source document, exemplars are also fed into the encoder. To make encourage the generated summary have similar architecture with these exemplars we link 
    sentences in the exemplars and sentences in generated summary by group tag embeddings, which means while decoding the $k$-th sentence of the target summary, it shares the same tag embedding of the $k$-th sentences in all these exemplars. }
\end{figure*}

\subsection{Summarizer}
To ensure that our framework can be orthogonal to the existing seq2seq models, we use two simple methods group alignment and ROUGE Credit. Figure~\ref{fig:architecture} shows the architecture of our \textsc{RetrievalSum} system, the encoder-decoder framework here can be instantiated with any popular pre-trained models without modifying internal structure. Specifically, we synthesize the document $X$ and the corresponding exemplars $\mathcal{E}$ in a single input sequence  $S$ = $(X, E_{1}, E_{2},\ldots, E_{|\mathcal{E}|})$. They are separated by the special tokens `[SEP]' and `[CLS]'. In the inference stage, we introduce the ROUGE credit method so that exemplars can guide the model to learn the summary format of the current corpus.


\subsubsection{Group Alignment}
Inspired by the success of contextualized rewriting  with group alignment~\cite{DBLP:journals/corr/abs-2102-00385}, we force the model to learn from the exemplars by aligning sentences. By aligning the sentences in the generated summary with the sentences in the exemplars, we expect that the model can learn the structural style of summary shared in the same domain.
Take scientific domain as an example, the structure of abstract usually follows the same paradigm: problem definitions, motivation, proposed methods, and experimental results.

As is shown in Figure~\ref{fig:architecture}, we add group tag embedding in both encoder and decoder.
For the $i$-th sentences in each exemplar, we set its group index as $G_i$ = $i$. In other words, we add a randomly initialized group tag embedding with index $i$ to the representation of each token in the $i$-th sentence.
Besides, we also introduce the group tag $G_0$ to tokens in the source document $X$ to distinguish $X$ from the the exemplars. We obtain the encoder output $\mathbf{h}_{enc}$ by:
\begin{equation}
\label{eq:enc}
\mathbf{h}_{enc} = \mathrm{Encoder}(S) +\mathrm{E}(G_i),
\end{equation}
where $\mathrm{E}(\cdot)$ is the group tag embedding layer.

In addition to the `[BOS]' and `[EOS]' tokens which are necessary for the generation model, token '[SEP]' is also inserted into the target summary sequence  $Y$  as a indicator of sentences boundaries. Similar to the group tag in the exemplars, we set the group index for the $i$-th sentences in the generated summary as $G_i$ = $i$.
We also add the document group tags $G_0$ to the tokens in the output summary. Thus we get the decoder ouput at step $k$ as:
\begin{align}
\label{eq:dec_out}
\mathbf{h}_{dec}^k &= \mathrm{Decoder}(\mathbf{h}_{enc}, \mathbf{h}_{dec\_in}^{<k}),\\
\mathbf{h}_{dec\_in} &= \mathrm{Emb}(\mathbf{t}) + \mathrm{E}(G_i) + \mathrm{E}(G_0),
\end{align}
where $\mathbf{t}$ is the target sequence, $\mathrm{Emb(\cdot)}$ is the token embedding layer and $\mathbf{h}_{dec\_in}$ represents the input embedding to the decoder at the current step.
Therefore, while generating the beginning/middle/end of the summary, the decoder can pay more attention to the corresponding part in the exemplars without losing the information in the source document.

Finally, our training object is to minimize the cross entropy loss:
\begin{align}
&\mathcal{L_{\theta}} =-\frac{1}{n}\sum_{k=1}^{n}\mathrm{log}P(t_k|\mathbf{t}_{<k},X,\mathcal{E},\theta),\\
&P(t_k|\mathbf{t}_{<k}, X,\mathcal{E},\theta) =\mathrm{softmax}(W\mathbf{h}_{dec}^k),
\end{align}
where $W \in \mathbb{R}^{vocab\times d}$ is a trainable parameter, we use weight sharing between $W$ and the token embedding matrix.

\subsubsection{ROUGE Credit}
In addition to the retrieved exemplars serving as input background knowledge in the training phase,  we also incorporate them in the inference stage to encourage the model to select the summary which has similar format with the retrieved exemplars.
In open-domain text generation, \citet{zou2021controllable} achieves substantial improvement via scoring the candidates produced by beam search conditioned on the given prompt. With the similar motivation,  we introduce ROUGE Credit~\cite{an2021enhancing} to encourage the model to keep paths with high relevance and style consistency continuously generating but dropping beams which  are irrevanlent with the exemplars.
While decoding the $k$-th token, we first select the exemplar $E_{best}$ which are paid the most attention by the decoder using $\operatorname{argmax}$ function over cross attention distribution on the '[CLS]' token inserted before each exemplars from the last encoder layer.
During the beam search, there are several candidate sequences $Cand$ = $(c_1, \ldots, c_{|Cand|})$ per time step, then we calcaute the ROUGE score between $E_{best}$ and each candidate sequence $c_i$ as:
\begin{align}
\label{eq:rougeCredit}
credit_{i} = {\rm ROUGE}(c_{i}, E_{best})*g(k)\\
g(k)=\left\{
\begin{array}{rcl}
exp(1- l_s/k)       &      & {k>l_s}\\
0 \qquad\qquad     &      & {k \leq l_s}\\
\end{array} \right.
\end{align}
where $g(k)$ is a weight function of the decoding step  $k$ and $l_s$ is a hyperparameter to control the start of ROUGE credit.
The score of the $i$-th candidate beam sequence $c_i$ is calculated by the sum of its average log likelihood and $credit_{i}$.

\section{Experiments}
 To validate the effectiveness of our system and get more convincing experimental results, we conduct experiments on seven mainstream datasets: BillSum (US congressional bills), AMI (meeting transcripts), Reddit (social media), PubMed (scientific papers), CNN/DailyMail (news) , XSum (news) and SSN (scientific papers). Details about these datasets see Appendix~\ref{sec:datasets}.
\subsection{Implementation Detail}
Due to the limitation of computation resources, both our retriever and summarizer are implemented by the base version of pre-trained models. We evaluate all the summarization systems with the standard ROUGE score \cite{lin2004rouge}.

\paragraph{Retriever} We use RoBERTa-base~\cite{liu2019roberta} as the backbone of our retriever. The non-shared MLP layer consists of 3 linear layers with hidden size 768, between which residual connection and dropout (ratio=0.1) are added. The number of heads is set to 16 and the temperature $\tau$ in contrastive loss is set to 0.1. We use Adam optimizer~\cite{kingma2014adam} with warming-up follow the setting in Transformer~\cite{vaswani2017attention} and the maximum learning rate is set to $10^{-4}$. We divide these datasets into two group according to the average summary length. For datasets (XSum, Reddit and CNNDM) with shorter summary, we select the top 5 exemplars from the retriever whose maximum length is truncated to 64. We choose the top 3 exemplars for the datasets (BillSum, SSN, PubMed and AMI) with longer summary and truncate them to 256. We train the retriever for 2 epochs where each step with a batch size of 16. The training process costs 16 hours on 4 GeForce RTX 3090 GPUs. 

\paragraph{Summarizer} The only additional parameter of our model is the group tag embedding matrix $W_{tag} \in \mathbb{R}^{N \times d}$ where $N$ is the maximum number of tags which is set to  and $d$ is the hidden size of the backbone model. We validate our system based on BertAbs and BART. For BertAbs, we use the default setting in~\citet{liu2019text}. For BART, we use an AdamW~\cite{loshchilov2018fixing} optimizer with maximum learning rate = $3 \times 10^{-5}$ and it decays linearly with training steps. By adding more randomly initialized position embedding, we break the maximum length limitation of BERT and BART. The maximum document input is set to 1024 and total input length of the exemplars is set to 768. We calculate ROUGE Credit every 6 steps. Due to the limitation of computation resources,  both the baseline models and the retrieval enhanced models are built on the base version of pre-trained models. We train the summarization model for 5 epochs with a batch size of 4. The training process costs 28 hours on 4 GeForce RTX 3090 GPUs. 

\subsection{Baseline Models}
We use the following  models for comparison. \textbf{BertAbs}~\cite{liu2019text} is an abstractive summarization system with pre-trained encoder BERT. \textbf{Template}~\cite{oya2014template} is a template-based method for meeting summarization which constructs templates using multi-sentence fusion algorithm.
\textbf{GSum}~\cite{dou2020gsum} is a general framework utilizing guide signals. During training it needs the oracle guidance and during inference it utilizes the guidance generated automatically. 
\textbf{BART}~\cite{lewis2020bart} is a denoising pre-trained seq2seq framework achieving state-of-the-arts in various generation tasks. Its base version has 140M parameters.
\textbf{PEGASUS}~\cite{zhang2020pegasus} is a model specially pre-trained for abstractive summarization whose base version  contains 220M parameters.

\renewcommand\arraystretch{1.0}
\tabcolsep0.10 in
\begin{table}[t]
\center \footnotesize

    \centering
    \begin{tabular}{lcccc}
        \toprule
        \textbf{Datasets}  & \textbf{Random} & \textbf{TF-IDF} & \textbf{Dense.} & \textbf{Oracle} \\
        \midrule
        Reddit & 6.77 & 9.66 & 12.35 & 23.76\\
        AMI & 31.31 & 34.80 & 36.85& 41.77\\
        CNNDM  &7.47 & 14.28 & 15.75 & 20.81\\
        XSum & 6.66 & 15.10 & 15.55 & 31.21\\
        BillSum  &15.49 & 26.14 & 26.21 & 34.82\\
        SSN  &15.43 & 20.02 & 20.82& 24.13\\
        pubMed  & 13.17 & 19.62 & 20.23 & 24.22 \\
        \bottomrule
    \end{tabular}
\caption{Average ROUGE score $\tilde{R}$ = (R-1 + R-2 + R-L)/3  between the exemplar and the gold summary. The higher the score, the more relevant the two texts are. \textbf{Random} means that we randomly select a summary from the training set as an exemplar. \textbf{TF-IDF} means we run TF-IDF algorithm to get the exemplars. \textbf{Dense.} indicates that the exemplar retrieved by our retriever. \textbf{Oracle} is the best exemplar in this training corpus.}
\label{tab:validate_exps}
\end{table}

\renewcommand\arraystretch{1.0}
\begin{table*}[t]
\center \footnotesize
\tabcolsep0.10 in
\begin{tabular}{lccccccccc}
\toprule
\multicolumn{1}{c}{\multirow{2}[1]{*}{\textbf{Model}}}  &
\multicolumn{3}{c}{\textbf{Reddit}} &
\multicolumn{3}{c}{\textbf{PubMed}} &
\multicolumn{3}{c}{\textbf{XSum}} \\

 & \textbf{R-1} & \textbf{R-2} & \textbf{R-L} &
\textbf{R-1} & \textbf{R-2} & \textbf{R-L} &
\textbf{R-1} & \textbf{R-2} & \textbf{R-L} \\

\cmidrule(lr){1-1} \cmidrule(lr){2-4} \cmidrule(lr){5-7} \cmidrule(lr){8-10}

Transformer & 15.89 & 1.94 & 12.22 & 33.94 & 7.95 & 19.02 & 30.83 & 10.83 & 24.41 \\
PEGASUS$^*$ & 24.36 & 6.09 & 18.75 & 39.98 & 15.15 & 25.23 & 39.79 & 16.58 & 31.70 \\
BertAbs$^\dag$ & 26.92 & 6.35 & 19.81 & 42.90 & 17.35 & 38.88 & 38.76 & 16.33 & 31.15 \\
BART & 31.00 & 10.12 & 25.34 & 43.60 & 17.01 & 39.34 & 41.05 & 18.55 & 33.40 \\
\cmidrule(lr){1-1} \cmidrule(lr){2-4} \cmidrule(lr){5-7} \cmidrule(lr){8-10}
\textsc{Retrieval}(BERT) & 28.58 & 6.71 & 22.60 &  44.39 & 17.34 & 40.14 & 39.34 & 16.68 & 31.38 \\
\textsc{Retrieval}(BART)& \textbf{32.38} & \textbf{10.20} & \textbf{25.96} & \textbf{45.21} & \textbf{17.64} & \textbf{40.65} & \textbf{41.82} & \textbf{18.92} & \textbf{33.93} \\
\quad- TF-IDF Retriever & 31.18 & 6.71 & 25.03 & 44.85 & 17.38 & 40.40 & 41.45 & 18.72 & 33.60 \\
\quad- Concatenate & 30.89 & 10.13 & 24.98 & 44.21 & 17.53 & 39.89 & 41.27 & 18.70 & 33.56 \\
\bottomrule
\end{tabular}
\caption{Results on test sets of Reddit, PubMed and XSum. Results with $^\dag$ are from \citet{dou2020gsum}, results with $^*$ are from \citet{zhang2020pegasus}. \textsc{Retrieval} (BERT) and \textsc{Retrieval} (BART) are our retrieval-based models with different backbone models. - TF-IDF Retriever means we replace the dense retriever with term-based system TF-IDF and - Concatenate means we directly concatenate these retrieved exemplars after the document input.}
\label{tab:reddit pubmed xsum}
\end{table*}

\subsection{Experimental Results and Analysis}
\paragraph{Quality of Retrieved Exemplars}
Finding high-quality and highly relevant exemplars is essential to the   generation stage. So we first evaluate the retrieved exemplars by adopting a heuristic approach: for each document in the test set, we calculate the average ROUGE score $\tilde{R}$ = (R-1 + R-2 + R-L)/3 between its retrieved exemplar and gold summary. The result is shown in Table~\ref{tab:validate_exps}.
We define {Oracle} as the best exemplar in the training corpus, which can be viewed as the upper bound of our retriever performance. {Random} baseline is also introduced as the lower bound of retriever performance and TF-IDF\footnote{\url{https://github.com/scikit-learn/scikit-learn}} is the term-based system. In general, the retriever proposed in this paper achieves better performance above all the baseline systems. On the two datasets with the shortest summaries, XSum and Reddit, the performance of retriever is still far from the upper bound but greatly outperform the term-based system. On other datasets such as AMI, SSN and PubMed, the performance of our retrieved exemplars is almost close to the Oracle exemplars.

\begin{figure}[t]
    \centering
    \includegraphics[width=0.9\linewidth]{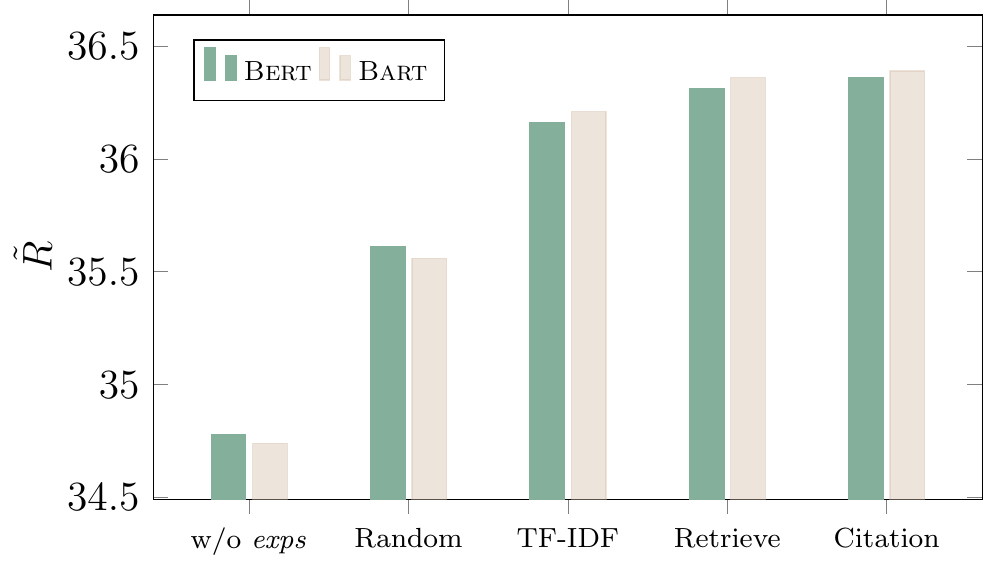}
    \caption{The impact of different exemplars sources on performance (SSN). X-axis denotes the sources and $\tilde{R}$ is the average of R-1, R-2 and R-L. Citation means the exemplars are retrieved based on the citation links of scientific papers.}
    \label{fig:methods}
\end{figure}

We further explore the  impact of different retrieval methods on model performance. The abstract section of papers in the same research community is a kind of easy-to-obtain but high-quality exemplars for scientific papers (denoted as \textit{citation} in Figure~\ref{fig:methods}). As can be seen from Figure~\ref{fig:methods}, for both BERT-based model and BART-based model, our retriever achieves comparable performance with citation. We also find that despite the ROUGE score between gold summaries and random exemplars is relatively low (\textbf{Random} baseline in Table~\ref{tab:validate_exps} with $\tilde{R} = 15.43$), but these exemplars still can contribute the performance improvement compared to the model purely relying on the source document. We think this shows that in the same domain, even bad exemplars can still provide help in writing format. As the quality of exemplars gradually improves, they can assist the model to a greater extent in understanding the content of the source document.

\renewcommand\arraystretch{1.0}
\tabcolsep 0.10in
\begin{table}[t]
\center \footnotesize
\begin{tabular}{lccc}
\toprule
\multicolumn{1}{c}{\textbf{Model}} & \textbf{R-1} & \textbf{R-2} & \textbf{R-L} \\
\midrule
\multicolumn{4}{c}{\textbf{AMI}} \\
\midrule
 Template & 31.50 & 6.80 & 16.99 \\
\textsc{BertAbs} & 45.74 & 14.21 & 42.09 \\

\textsc{Retrieval} (\textsc{Bert}) & 49.88\textbf{\tiny{$\uparrow$\textcolor{red}{4.14}}} & 14.96 & 47.19 \\
BART & 46.20 & 16.73 & 44.45 \\
\textsc{Retrieval} (\textsc{Bart}) & \textbf{50.86\tiny{$\uparrow$\textcolor{red}{4.66}}} & \textbf{16.76} & \textbf{48.45}  \\
\midrule
\multicolumn{4}{c}{\textbf{BillSum}} \\
\midrule
PEGASUS $^\dag$ & 51.42 & 29.68 & 37.78\\
\textsc{BertAbs} & 50.77 & 29.75 & 47.11 \\
\textsc{Retrieval} (BERT) & 53.29\textbf{\tiny{$\uparrow$\textcolor{red}{2.52}}} & 31.10 & 49.54 \\
BART & 51.80 & 33.05 & 47.72 \\
\textsc{Retrieval} (BART) & \textbf{56.26 \tiny{$\uparrow$\textcolor{red}{4.46}}} & \textbf{34.90} & \textbf{52.51}  \\
\bottomrule
\end{tabular}
\caption{Results on the test set of AMI (meeting transcripts) and BillSum (US congressional bills). Results with $^\dag$ are from \citet{zhang2020pegasus}.} 
\label{tab:billsum ami ssn}
\end{table}

\begin{figure}[t]
    \centering
    \includegraphics[width=0.85\linewidth]{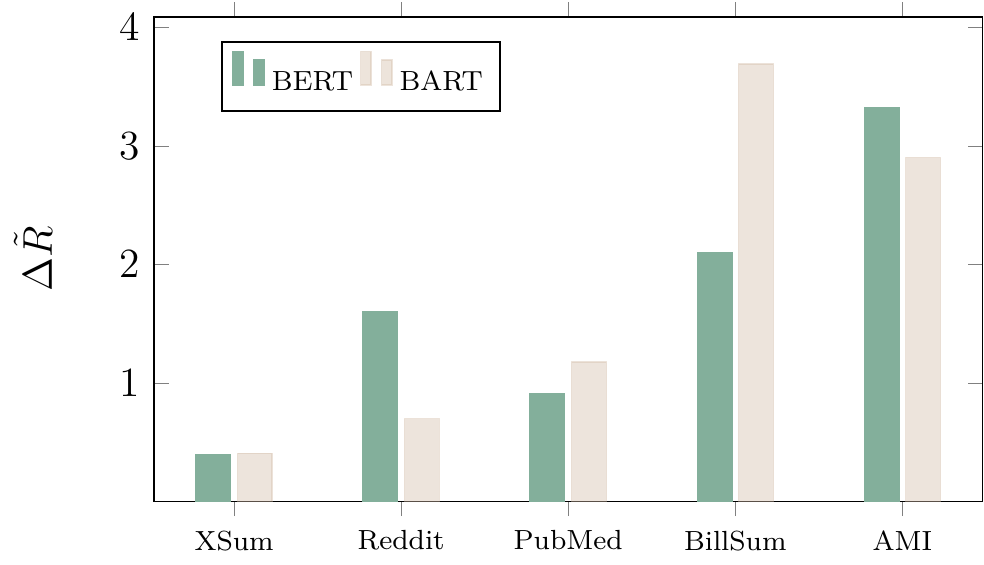}
    \caption{Comparison of the gain from exemplars across multiple  domains. X-axis denotes datasets from various domains: \textbf{\textit{scientic, bill, meeting, news and  social media}}  and $\Delta \tilde{R}$ is the performance gain in $\tilde{R}$}
    \label{fig:datasets}
\end{figure}
\paragraph{Evaluation of Generated Summary}
Table~\ref{tab:reddit pubmed xsum} shows our results on Reddit, PubMed and XSum. Generally, the performance of different baseline models can be significantly boosted with retrieval. For BERT-based models and BART-based models, our model has an average improvement of over 1.0 R-L score.  We also compare our method with simply concatenating the retrieved exemplars after the source document (denote as -Concatenate in Table~\ref{tab:reddit pubmed xsum}). For scientific papers which strongly need background knowledge, directly concatenating the retrieved summaries can lead a big jump in performance.  Group alignment and ROUGE credit can further enhanced the utilization of the structure information of retrieved exemplars and lead to better results.    

In Table~\ref{tab:billsum ami ssn}, we list the results on BillSum and AMI where target summaries are written with an obvious domain-specific writing format. On these datasets, our approach achieve very surprising results: we improve the general-purpose summarization model BART by more than \textbf{4.4} R-1 score and \textbf{4.0} R-L score.
Figure~\ref{fig:datasets} shows the performance gain of our BART-based models in different domains. On PubMed (Scientific papers), BillSum (Congressional bills), Reddit(social media) and AMI (meetings), our retriever can obtain more improvements than the other domains. For news , the improvements are relatively small due to the lack of close connections between samples, in other words thus bring difficulties for retrieving high-quality exemplars and their writing style are more flexible which may be unhelpful to guide other samples. 

\renewcommand\arraystretch{1.0}
\begin{table}[t]
       \small
  \centering \tabcolsep0.04 in
  \begin{tabular}{@{}llrrr@{}}
  \toprule
   \bf Model &  \bf  & \bf R-1 &\bf R-2 &\bf R-L \\
  \midrule
\multicolumn{2}{l}{Copy from TF-IDF} &20.66 & 3.18 & 19.00  \\
\multicolumn{2}{l}{Copy from Elastic Search} &20.81 & 3.22 & 19.23  \\
\multicolumn{2}{l}{Copy from Dense Retriever} & 23.04 & 3.25 & 20.96 \\
\multicolumn{2}{l}{Copy from Oracle} & 30.11& 5.44& 26.89 \\
\multicolumn{2}{l}{BertAbs} & 41.58 & 18.99 & 38.56 \\
\multicolumn{2}{l}{BART} & 42.93 & 20.01 & 39.56 \\
  \midrule
     \multirow{2}{*}{GSum (BERT)$^*$} &  Elastic Search &40.88 & 18.24 & 37.99 \\
    & Oracle  & 43.69 & 20.53 & 40.71 \\
    \cmidrule(l){1-1} \cmidrule(lr){2-2} \cmidrule(lr){3-5} 
    \multirow{4}{*}{\textsc{Retrieval} ({BERT})}  & TF-IDF & 41.11 & 18.29 & 38.14 \\
    & Dense Retriever& 41.64 & 18.69 & 38.86  \\
    & Elastic Search & 41.15 & 18.28 & 38.20\\
    & Oracle & 46.80 & 22.66 & 43.83 \\
    \cmidrule(lr){1-1} \cmidrule(lr){2-2} \cmidrule(lr){3-5} 

    \multirow{2}{*}{\textsc{Retrieval} ({BART})} & Dense Retriever & 42.91 & 19.70 & 40.11 \\
     & Oracle & 48.10 & 23.70 & 45.09 \\
    \bottomrule
    \end{tabular}
    \caption{ \label{tab:cnndm} Comparison with GSum which uses BertAbs as the backbone and guidance from retrieved summaries on CNNDM dataset. Copy from TF-IDF/ Elastic Search/ Dense Retriever/ Oracle mean we random copy an exemplar from the corresponding candidate set as the output summary. Results with $^*$ are reported in \citet{dou2020gsum}.} 
  \end{table}

\paragraph{Ability to Learn from Exemplars}
\citet{dou2020gsum} report the performance of BertAbs with the guidance of summaries retrieved via lexical matching algorithm Elastic Search\footnote{\url{https://github.com/elastic/elasticsearch}} (in Table~\ref{tab:cnndm}) which has the similar input with our model. We experiment with various retriever: TF-IDF, Elastic Search, our dense retriever. In addition to the results obtained by automatic retrieval, we also compare the Oracle performance with their model. As is shown in Table~\ref{tab:cnndm}, for model with the same backbone model BertAbs, our model beats GSum by 0.76 points R-1 and 0.87 points in R-L. If we initialize the retriever as Elastic Search and Oracle, the two systems have almost the same input. We can see that our model can achieve the advantage of  \textbf{3.11} R-L scores \textbf{3.12} R-L scores over GSum when given higher quality exemplars. This indicates that the architecture of our model can benefit more from retrieved exemplars than previous methods.

\paragraph{Human Evaluation}
\renewcommand\arraystretch{0.8}
\tabcolsep 0.08in
\begin{table}[t]
    \centering \footnotesize
    \begin{tabular}{lcccc}
        \toprule
        \textbf{Systems}  & \textbf{Fluency.} & \textbf{Info.} & \textbf{Faith.} & \textbf{Style.} \\
        \midrule
        \textsc{BertAbs} & 4.36 & 3.84 & 4.05 & 4.13\\
        \textsc{Retrieval}-\textsc{Bert} & 4.50 & 4.02 & 3.95& 4.54\\
        \textsc{Bart} & 4.52 & 3.98 & 4.35 & 4.46\\
        \textsc{Retrieval}-\textsc{Bart} & 4.63 & 4.15 & 4.37 & 4.72\\
        Ground-Truth & 4.80 & 4.68 & 4.86 & 4.92\\
        \bottomrule
    \end{tabular}
\caption{ Results of human evaluation on fluency, informativeness, faithfulness and writing style  on the test set of BillSum.}
\label{tab:human_evaluation}
\end{table}
 To check are the output summaries written in a correct style and what leads such a surprising improvement in BillSum,  we conduct a human evaluation on the test set of BillSum. We randomly select 50 articles from the test set,  and each articles has 5 candidate summaries where 4 from automatic systems and 1 from human written. We ask the participants to score the fluency, informativeness, faithfulness of the candidate summaries from 1.0 to 5.0 (the higher the better). To confirm that our model has the ability to capture the writing style of the given corpus,  annotators are asked to score the generated summary based on some high-quality examples selected by us and their common sense of the style of congressional bills. Each sample is scored by 3 independent annotators and we report the average score.  As shown in Table~\ref{tab:human_evaluation}, the output of retrieval enhanced model has great advantages in terms of informativeness and fluency.    
 We do not observe a significant increase or drop in faithfulness. The style of our generation results obtains much improved(from 4.46 to 4.72) and only lags behind human written summaries by 0.2 points.  An \textbf{interesting finding} is that current state-of-the-art model are not good at generating long text but it can be alleviated if give some `prompts' which are always in front of the key idea of the source document. For instance, if the model are reminded to start with \textit{`amends the internal revenue code to allow...'}(Bills) or \textit{`in this paper, we propose...'}(papers), it will force the decoder catch the key point and lead to a more informative summary. High-quality retrieved exemplars always contains such prompts and the powerful PTMs are able to learn copying useful prompts to form a well-formatted summary while baseline models are more likely to deviate from the content without reminded by retrieved exemplars.

\section{Conclusion}
We propose a novel retrieval enhanced framework \textsc{RetrievalSum} for abstractive summarization. Instead of generating a summary merely depending on the source document, we firstly retrieve several semantically-similar exemplars and then utilizes these exemplars as guidance for better understanding the source document and learning domain-specific writing style. Experiments on seven datasets show the effectiveness of our model.

\bibliography{anthology,custom}
\bibliographystyle{acl_natbib}

\appendix
\begin{table*}[t]
  \centering  \tabcolsep0.10 in
    \begin{tabular}{lcccccc}
    \toprule
    \multicolumn{1}{c}{\multirow{2}[1]{*}{\textbf{Datasets}}}  &
    \multicolumn{3}{c}{BART} &
    \multicolumn{3}{c}{BART (Oracle)} \\
    & \textbf{R-1} & \textbf{R-2} & \textbf{R-L} & \textbf{R-1} & \textbf{R-2} & \textbf{R-L}\\
    \cmidrule(lr){1-1} \cmidrule(lr){2-4} \cmidrule(lr){5-7} 
    CNNDM & 42.93 & {20.01} & {39.56} & 48.10 & 23.70 & 45.09 \\
    Reddit & 31.00 & 10.12 & 25.34 & 40.63 & 15.61 & 32.14 \\
    AMI & 46.20 & 16.73 & 44.45 & 53.62 & 18.82 & 50.86 \\
    XSum & 41.05 & 18.55 & 33.40 & 47.44 & 23.82 & 38.71 \\
    BillSum & 51.80 & 32.99 & 51.46 & 59.44 & 37.49 & 55.47 \\
    PubMed & 43.87 & 16.59 & 39.62 & 47.09 & 18.94 & 42.29 \\
    \bottomrule
    \end{tabular}
\caption{Oracle experiment results on the test set of CNNDM, Reddit, AMI, XSum, BillSum and PubMed with baseline model as BART-base (140M). Results exceeding SOTA are in bold.}
\label{tab:oracle_exps}
\end{table*}

\section{Datasets}\label{sec:datasets}
\paragraph{CNN/DailyMail}~\cite{hermann2015teaching} is a commonly used summarization dataset modified by \citet{nallapati2016abstractive}, which contains news articles and
associated highlights as summaries. In this paper, we use the non-anonymized version follwing~\cite{see2017get}. 

\paragraph{PubMed}~\cite{cohan2018discourse} is collected from scientific papers and therefore consists of long documents  containing 133k samples  In our experiment, we truncate the input length to 1024. 

\paragraph{BillSum}~\cite{DBLP:journals/corr/abs-1910-00523} contains 23k US Congressional bills and human-written reference summaries
from the 103rd-115th (1993-2018) sessions of Congress. The California test set which is out-of-distribution is not used. We use its tokenized version.

\paragraph{XSum}~\cite{narayan2018don} is a one-sentence summary dataset to answer the question ``What is the article about?''  consists of 227k BBC articles from 2010 to 2017 covering a wide variety of subjects along. All summaries are professionally written, typically by the authors of documents in this dataset.

\paragraph{SSN}~\cite{an2021enhancing} is a scientific papers dataset with 144k samples. Different from arXiv, SSN  provide citation relations between these papers. Most of the papers belongs to computer science, physics and maths. We use their inductive dataset division. 

\paragraph{Reddit}~\cite{kim2019abstractive} is a highly abstractive dataset collected from social media platform. We only use the TIFU-long version of Reddit, which regards the body text of a post as the document and the TL;DR as the summary.

\paragraph{AMI}~\cite{kim2019abstractive} is a meeting summarization dataset consists of 137 meetings transcripts about the design of a remote control. Gold referece is annotated by human. We follow \citet{shang-etal-2018-unsupervised} for the dataset division.

\section{Automatic Factual Consistency Evaluation}
\begin{figure}[t]
    \centering
    \includegraphics[width=0.95\linewidth]{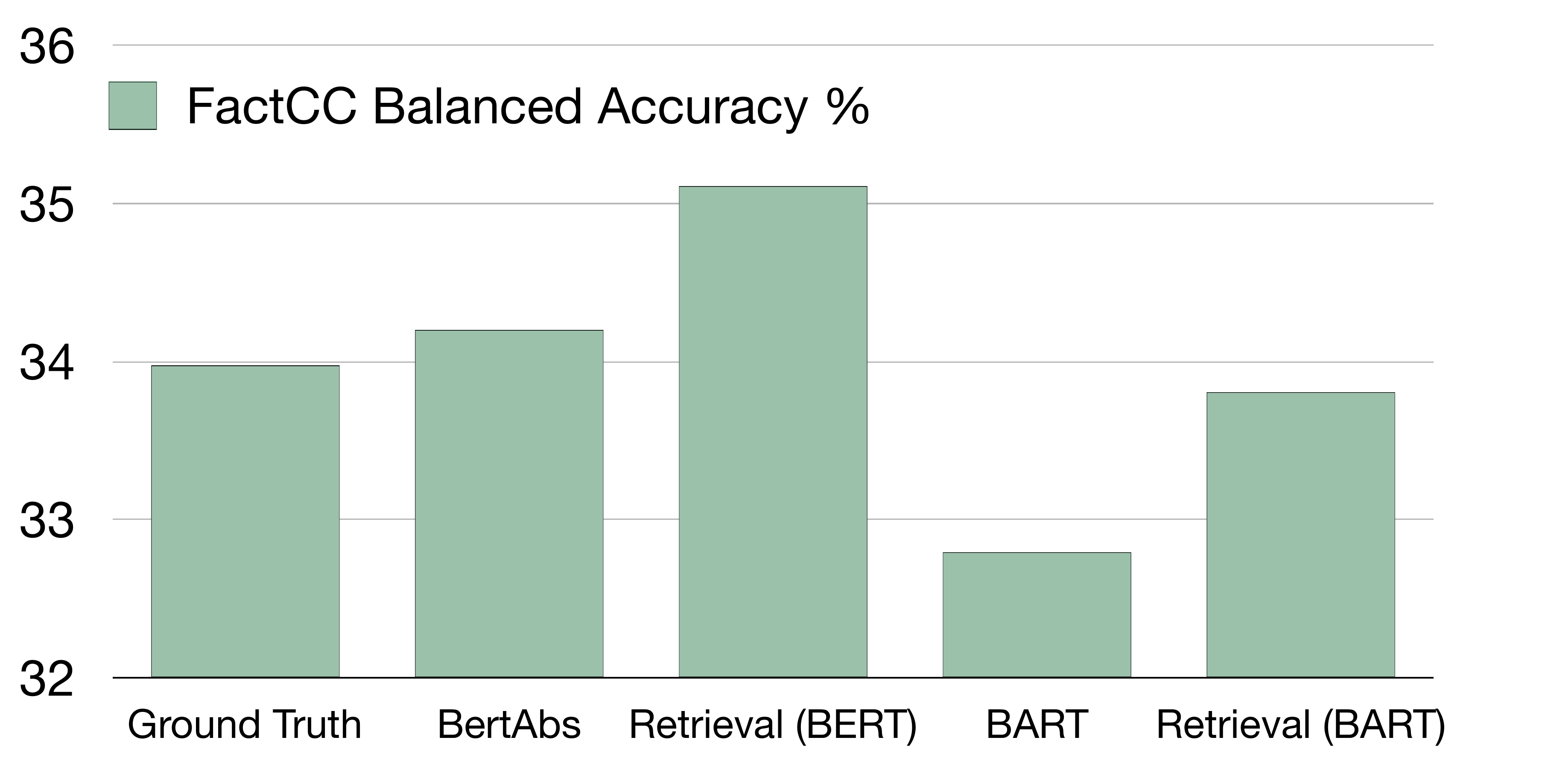}
    \caption{The balanced accuracy output by FactCC}
    \label{fig:factcc}
\end{figure}

In addition to human evaluation, we also try the automatic evaluation tool FactCC~\cite{kryscinskiFactCC2019} shown in Figure~\ref{fig:factcc}. The non-retrieval model \textsc{BertAbs} model achieve the best performance. However, these results are completely inconsistent with our human evaluation meanwhile the FactCC tool will give the gold reference a very low accuracy (32.9\%). Considering our model is optimized towards the gold reference, the factCC score might
not be a good indicator of whether there are factual errors in a generated summary.

\section{Ablation Study}
\begin{table}[t]
    \centering
    \begin{tabular}{lccc}
        \toprule
        \textbf{Model}  & \textbf{R-1} & \textbf{R-2} & \textbf{R-L} \\
        \midrule
        \textsc{Retrieval}(BART) & \textbf{56.26} & \textbf{34.90} & \textbf{52.51} \\
        \quad - Group Alignment &55.25 & 34.21 & 51.95\\
        \quad - ROUGE Credit &53.23 & 34.04 &48.29\\
        \quad - Retrieval & 51.80 & 33.05 & 46.72\\
        \bottomrule
    \end{tabular}
    \caption{Ablation study of the \textsc{RetrievalSum}. ’-’ means we remove the module from the original model.} 
    \label{tab:Ablation}
\end{table}

To gain a better understanding of the contribution of the components in our summarizer. We conduct comprehensive ablation studies on BillSum based on the state-of-the-art pretrained model  BART\cite{lewis2020bart}. We respectively remove  Group Alignment, ROUGE Credit and the retrieved exemplars. Results are reflected in Table~\ref{tab:Ablation}.
\section{Oracle Experiment}
We also perform an Oracle experiment (shown in Table~\ref{tab:oracle_exps}) where we directly use the Oracle exemplars to help the generation. Results shows that even using the \textbf{base} version of pre-trained model as the backbone model of our framework whose number of parameters is only 140M , \textsc{RetrievalSum} with Oracle exemplar  significantly outperform the non-retrieval version and obtain the \textbf{state-of-the-art} performance on multiple datasets. The results illustrates that the design of our summarizer can be adapted to exemplars of various quality and better retrieval methods can further help the generation model.  It also reveals that the current retriever still has untapped potential, and further research is urgently needed.
\begin{table*}[t!]
    \scriptsize
    \centering
    \extrarowheight=\aboverulesep
    \addtolength{\extrarowheight}{\belowrulesep}
    \aboverulesep=1pt
    \belowrulesep=1pt
    \begin{tabular}{@{}c  p{0.82\textwidth}}
     \toprule
     \multicolumn{1}{c}{\bf Document} & < section - header > refundable credit for child disability education and training expenses .  subpart c of part iv of subchapter a of chapter 1 of the internal revenue code of 1986 is amended by redesignating section 35 as section 36 and inserting after section 34 the following new section : ` ` section 35 .  child disability education and training expenses .  allowance of credit .  in the case of an individual , there shall be allowed as a credit against the tax imposed by this subtitle for the taxable year an amount equal to the amount paid or incurred by the taxpayer during the taxable year for qualified child disability expenses .  maximum dollar amount .  the amount allowed as a credit under subsection ( a ) to the taxpayer for the taxable year shall not exceed \$ 3 , 000 .  limitation based on adjusted gross income .  the amount of the credit allowable under subsection ( a ) ( after application of paragraph ( 1 ) ) shall be reduced by \$ 500 for each \$ 1 , 000 by which the taxpayer ' s modified adjusted gross income exceeds \$ 150 , 000 .  modified adjusted gross income .  for purposes of subparagraph ( a ) , the term ` modified adjusted gross income ' means adjusted gross income increased by any amount excluded from gross income under section 911 , 931 , or 933 .  in the case of any taxable year beginning in a calendar year after 2002 , the \$ 150 , 000 amount under subparagraph shall be increased by an amount equal to such dollar amount , multiplied by the cost - of - living adjustment determined under section 1 ( f ) ( 3 ) for the  if any amount after adjustment under clause ( i ) is not a multiple of \$ 1 , 000 , such amount shall be rounded to the next lower multiple of \$ 1 , 000 .  qualified child disability expenses .  for purposes of this section in general .  the term ` qualified child disability expenses ' means amounts paid for services and equipment related to education and training of a qualified child of the taxpayer in connection with a developmental disability of such child , including behavioral therapy , speech therapy , occupational therapy , physical therapy ,  the term ` developmental disability '...
      \\\cmidrule{1-2}
    \multicolumn{1}{c}{\bf Exemplar 1} &  foster care tax credit act this bill \textbf{amends the internal revenue code to allow a partially refundable tax credit for each qualifying foster child who resides in the home of an eligible taxpayer for at least one calendar month during the taxable year} . a quot , qualifying foster childquot ... no credit is allowed if the identification number of either the taxpayer or the qualifying child was issued after the due date for filing the return for the taxable year . the bill denies the tax credit to certain taxpayers who have made prior fraudulent or reckless claims for the credit within specified disallowance periods . the department of health and human services must identify provisions in the internal revenue code that can benefit foster families and increase outreach efforts to inform state and indian tribal foster care agencies and foster families about such provisions . \\
    \multicolumn{1}{c}{\bf Exemplar 2} &  comprehensive charity reform act - \textbf{amends the internal revenue code to allow an individual a tax credit not exceeding \$500 for qualified charitable contributions paid to certain private charities providing assistance to the poor} . sets forth provisions providing for the coordination of the credit with deductions allowable for charitable contributions . allows an individual who does not itemize deductions for the taxable year a direct charitable deduction in the amount allowable for qualified charitable organizations . exempts the charitable contribution deduction from the overall limitation on itemized deductions . allows the taxpayer to elect to treat any charitable contribution made before the time prescribed by law for filing of the return for the taxable year as being made on the last day of such taxable year and makes such an election irrevocable .\\
    \multicolumn{1}{c}{\bf Exemplar 3} &  \textbf{amends the internal revenue code to allow an income and excise tax credit} for the production of algae derived fuel and fuel mixtures . defines algae derived fuel as a liquid hydrocarbon product that is derived from the biomass of algal organisms . terminates such credits after 2012 .
 \\\cmidrule{1-2}
 \multicolumn{1}{c}{\bf Human written.} 
     & \textbf{amends the internal revenue code to allow a tax credit for up to \$3,000 of the amount paid or incurred by the taxpayer} during the taxable year for qualified child disability expenses for services and equipment related to education and training of a child of the taxpayer with a developmental disability , including behavioral therapy , speech therapy , occupational therapy , physical therapy , auditory therapy , assistive communication technology , and other related services .phases out the credit by \$500 increments per \$1,000 of modified adjusted gross income over \$150,000.
 \\\cmidrule{1-2}
    \multicolumn{1}{c}{\cellcolor{gray!25}\bf \textsc{RetrievalSum}} 
     & \cellcolor{gray!25} \textbf{amends the internal revenue code to allow an individual a refundable tax credit for qualified child disability education and training expenses.} limits the amount of such credit to \$3,000 . reduces such amount by \$500 for each \$1,000 by which the taxpayer's modified adjusted gross income exceeds \$150,000 , and provides a limitation based on adjusted gross incomes. provides for an inflation adjustment to such amount.allows a deduction for expenses relating to the education of a qualified child of the taxpayer in connection with a developmental disability of such child. directs the secretary of the treasury to report to the congress on the effectiveness of the credit in this act in the recruitment and training of qualified child victims in the united states .
 \\\cmidrule{1-2}
 \multicolumn{1}{c}{\bf BART} 
 & for the purposes of this section in general, qualifying child disability expenditures. \textbf{limits the amount of such credit to \$3 , 000 for each \$1 , 000 by which the taxpayer ' s modified adjusted gross income exceeds \$ 150,000.} provides for an inflation adjustment based on adjusted gross incomes and a limitation based on the gross income of the taxpayer and the taxpayer identification number of such a child on the return of the tax for the taxable year. prohibits the credit from being allowed for any expense to the extent that funds for such expenses are received under any federal , state , or local program  . requires married couples to file joint returns . allows a married couple to file a joint tax return.
 \\
  \bottomrule
\end{tabular}
\caption{ A Case study from the test set of BillSum. Text in bold reflects the main idea of the bill. We can see that with the help of exemplar from the same domain, the quality of generation result is greatly boosted.}
\label{tab:example}
\end{table*}

\section{Case study}
Table~\ref{tab:example} gives a case study from the test set of BillSum. The source document is about tax credit for  qualified child disability education and the three retrieved exemplars  are summaries of tax credit for other events. Summaries of these bills regarding tot tax credit  often share  similar writing style or format. We show that taking these retrieved exemplars as reference during the decoding process leads a more well-formatted and informative summary.

\end{document}